\documentclass[journal]{IEEEtran}
\usepackage{color}
\usepackage{changes}

\usepackage{tabularx}
\usepackage{xparse}
\usepackage{amsmath}
\usepackage{amsthm}
\usepackage{amssymb}
\usepackage{xspace}
\usepackage{relsize}
\usepackage{graphicx}
\usepackage{multirow}
\usepackage{url}
\usepackage{comment}
\usepackage{multirow}
\usepackage{booktabs}
\usepackage{bbding}

\newcommand{\ie}{\textit{i.e.}\xspace}
\newcommand{\eg}{\textit{e.g.}\xspace}
\newcommand{\etal}{\textit{et al.}\xspace}

\ifCLASSINFOpdf
\else
\fi
\definechangesauthor[name={weiz}, color=blue]{wz}
\begin{document}
\title{Affinity Space Adaptation for Semantic Segmentation Across Domains}
\author{Wei Zhou, Yukang Wang, Jiajia Chu, Jiehua Yang, Xiang Bai, Yongchao Xu
% \author{Wei Zhou,~\IEEEmembership{Member,~IEEE,}
%         ,~\IEEEmembership{Fellow,~OSA,}
%         and~Jane~Doe,~\IEEEmembership{Life~Fellow,~IEEE}% <-this % stops a space
  \thanks{W. Zhou, and Y. Xu are with the School of Computer Science, Wuhan University, Wuhan, 430072, China. W. Zhou, Y. Wang, J. Chu, J. Yang, X. Bai, and Y. Xu are with the School of Electronic Information and Communications, Huazhong University of Science and Technology, Wuhan 430074, China (e-mail:weizhou@hust.edu.cn; wangyk@hust.edu.cn; jiajia\_chuvip@hust.edu.cn; jiehuayang@hust.edu.cn; xbai@hust.edu.cn; yongchao.xu@whu.edu.cn). \textit{(corresponding author: Yongchao Xu.)}}}


\markboth{Journal of \LaTeX\ Class Files,~Vol.~XX, No.~XX, January~2020}%
{Shell \MakeLowercase{\textit{et al.}}: Bare Demo of IEEEtran.cls for IEEE Journals}



\maketitle

\begin{abstract}
Semantic segmentation with dense pixel-wise annotation has achieved excellent performance thanks to deep learning. However, the generalization of semantic segmentation in the wild remains challenging.
In this paper, we address the problem of unsupervised domain adaptation (UDA) in semantic segmentation.
Motivated by the fact that source and target domain have invariant semantic structures, we propose to exploit such invariance across domains by leveraging co-occurring patterns between pairwise pixels in the output of structured semantic segmentation.
This is different from most existing approaches that attempt to adapt domains based on individual pixel-wise information in image, feature, or output level.
Specifically, we perform domain adaptation on the affinity relationship between adjacent pixels termed affinity space of source and target domain.
To this end, we develop two affinity space adaptation strategies: affinity space cleaning and adversarial affinity space alignment.
Extensive experiments demonstrate that the proposed method achieves superior performance against some state-of-the-art methods on several challenging benchmarks for semantic segmentation across domains.
The code is available at \url{https://github.com/idealwei/ASANet}.
\end{abstract}

\begin{IEEEkeywords}
Unsupervised domain adaptation (UDA), semantic segmentation, affinity relationship.
\end{IEEEkeywords}

% \fi
\IEEEpeerreviewmaketitle

\section{Introduction}

\IEEEPARstart{F}{ollowing} the pioneer fully convolutional network (FCN)~\cite{long2015fcn} for semantic segmentation, recent methods~\cite{zheng2015conditional,zhao2017pspnet,chen2018deeplab,zhang2018cenet,zhao2018psanet,ke2018aaf} have achieved remarkable performances in semantic segmentation using dense pixel-wise annotation. They usually endeavor to boost segmentation accuracy by enlarging receptive fields while preserving fine detail information~\cite{chen2018deeplab}, making use of context information~\cite{zhao2017pspnet,zhang2018cenet,zhen2019fullydense}, capturing long range dependency via attention mechanism~\cite{zhao2018psanet}, or incorporating structural reasoning via pairwise affinity~\cite{zheng2015conditional,ke2018aaf}. Despite these efforts, it is common that a segmentation model trained on a specific domain fails to generalize well on a new one. This is because there exists domain shift between source training images and target testing images. To overcome this, one usually resorts to large amount of pixel-wise labeled target data, which is expensive and tedious to collect. Therefore, semantic segmentation in the wild remains challenging.

 \begin{figure}
    \centering
    \includegraphics[width=1.0\linewidth]{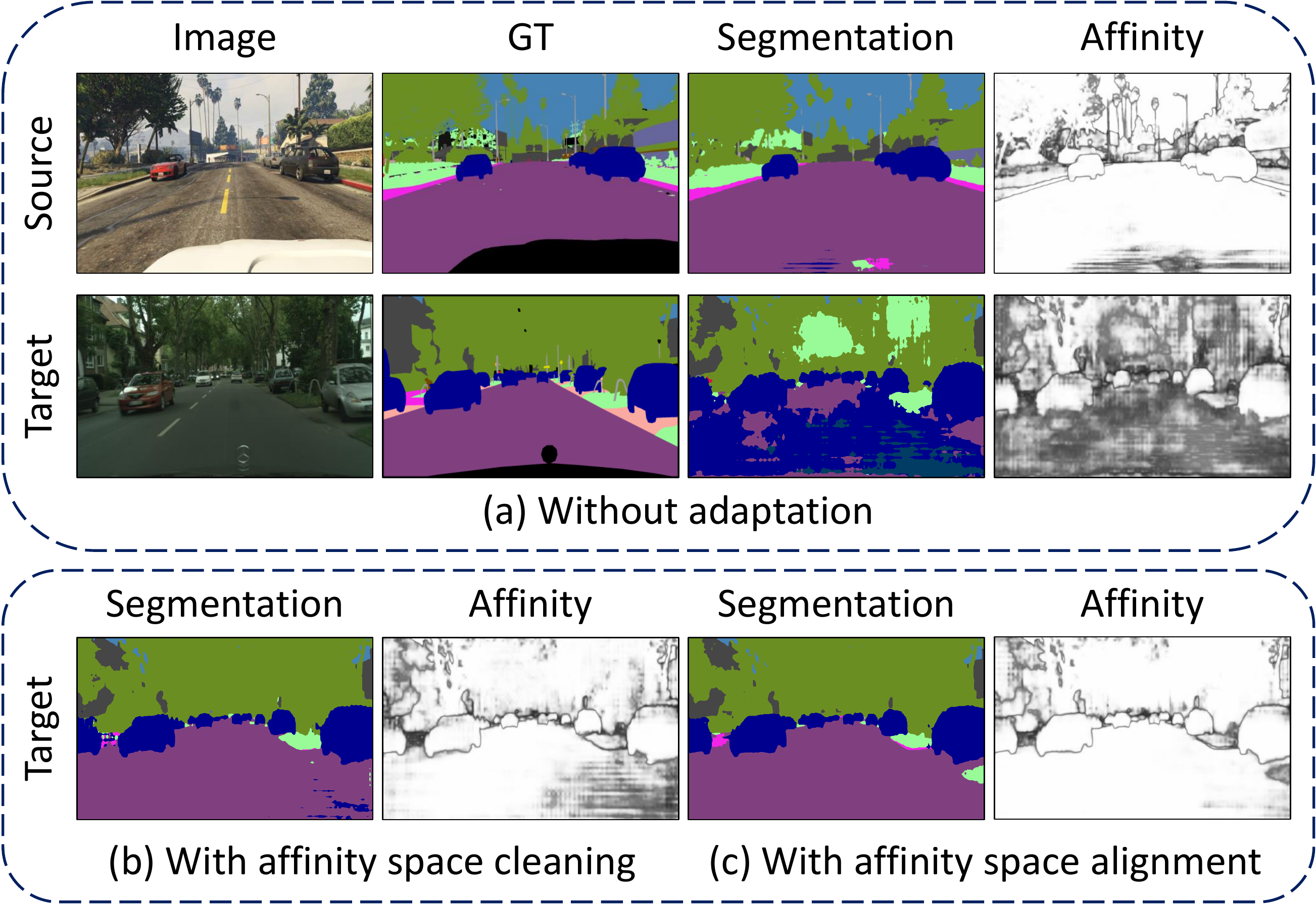}
    \caption{Proposed affinity space adaptation motivated by similar output structure across domains. We propose to match the affinity space revealing output structure between source and target domain. (a) shows the segmentation results of source and target domain without adaptation. (b) and (c) show the corresponding segmentation results with the proposed \textrm{affinity space cleaning (ASC)} and \textrm{affinity space alignment (ASA)} strategy for affinity space adaptation. For affinity visualization, we show the average Cosine similarity between predictions of each pixel and all its adjacent pixels.}
    \label{fig:introaff}
\end{figure}

Unsupervised domain adaptation (UDA) approaches have been recently studied to address the above issue, where only annotations of source domain are provided, but not for the target domain. Such UDA problem has been extensively exploited for image classification, where most prior works attempt to match source and target distributions to learn domain-invariant features by adversarial learning~\cite{goodfellow2014gan}. Similarly, in the field of semantic segmentation, most recent approaches seek to bridge domain gaps by minimizing the difference between the distributions of image style~\cite{hoffman2017cycada}, intermediate features~\cite{hoffman2016fcns,sankaranarayanan2018lsd,zhang2018fcan}, or network output space~\cite{tsai2018adaptsegnet}. These existing methods suggest that alignment in image, feature, and output level plays an important role in cross-domain semantic segmentation.

Considering that semantic segmentation is a structured prediction problem, which is robust across domains, it is reasonable to hypothesize that structural knowledge would be beneficial for semantic segmentation across domains.
The affinity relationship between output of neighboring pixels contains rich information about spatial structure and local context, which reveals the structural knowledge.
Therefore, we introduce the concept of affinity space built upon the affinity relationship between prediction output of each pixel and all its adjacent pixels. We aim to conduct domain adaptation for semantic segmentation on top of this concept.
Specifically, we explore two implementations of such concept related to two schemes in different perspectives to achieve such affinity space adaptation.
Firstly, we design an affinity loss that forces high affinity everywhere in the affinity space of both target and source images in addition to segmentation loss on the source domain. Since an image (in both source and target domain) is a priori composed of some continuous semantic regions. Adjacent pixels belonging to the same semantics are dominant over adjacent pixels lying on two different semantic regions. Therefore, such designed affinity loss acts as affinity space cleaning that regularizes the affinity space and results in an affinity space with high value almost everywhere for the target domain, approaching that of source domain. Indeed, as depicted in Fig.~\ref{fig:introaff}(a), without adaptation, the affinity space of the target domain is not clean in the sense that there are many low affinities. The proposed affinity space cleaning (see Fig.~\ref{fig:introaff}(b)) yields clean and similar affinity maps for the source and target domain.  Secondly, we also investigate adversarial training to directly align affinity space distribution of the target domain and that of the source domain. As depicted in Fig.~\ref{fig:introaff}(c), adversarial affinity space adaptation also effectively aligns the affinity space.

In summary, the main contributions of this paper are three-fold:
1) We introduce the concept of affinity space that highlights structure through leveraging co-occurring output patterns between neighboring pixels for domain adaptation in semantic segmentation;
2) We propose two effective schemes for affinity space adaptation: affinity space cleaning via affinity loss and adversarial affinity space alignment;
3) The proposed affinity space adaptation achieves superior performance over some state-of-the-art methods on several challenging benchmarks for semantic segmentation across domains.

The rest of this paper is organized as follows. We shortly review some related works in Section~\ref{sec:relatedwork}, followed by the proposed method in Section~\ref{sec:approach}. Section~\ref{sec:experiments} presents extensive experimental results. Finally, we conclude and give some perspectives in Section~\ref{sec:conclusion}.

\section{Related Work}
\label{sec:relatedwork}

We first review some representative works on semantic segmentation in Section~\ref{subsec:ss}, followed by some related works that also leverage pairwise affinity in Section~\ref{subsec:pa}. We then shortly review some recent domain adaptation methods for semantic segmentation across domains in Section~\ref{subsec:da}. The comparison of the proposed method with some related works is given in Section~\ref{subsec:comparison}.

\subsection{Semantic segmentation.}
\label{subsec:ss}
% definition
Semantic segmentation is a fundamental task in computer vision. The goal is to assign a category label to each pixel of the image. Driven by the power of deep neural networks, semantic segmentation has achieved great progress since pioneer works: FCN~\cite{long2015fcn} and U-Net~\cite{ronneberger9351unet}.

Numerous methods have then been proposed to boost segmentation performance. For example, Deeplab~\cite{chen2018deeplab,deeplabv3plus2018} shows that enlarging receptive fields through dilated convolutions leads to significant performance gain. Exploiting context information through atrous spatial pyramid pooling (ASPP)~\cite{chen2018deeplab,deeplabv3plus2018}, pyramid spatial pooling (PSP)~\cite{zhao2017pspnet} or context encoding~\cite{zhang2018cenet} also improve the segmentation accuracy. Some other methods~\cite{zheng2015conditional, liu2017spn, ke2018aaf} attempt to incorporate pairwise relation to provide structured reasoning that helps to alleviate inconsistency issue in semantic segmentation. Recently, DANet~\cite{fu2019danet} and OCNet~\cite{yuan2018ocnet} propose to utilize self-attention mechanism to capture long range dependency and achieve superior performances.

These fully supervised methods rely on large amount of pixel-wise annotated data. However, it is claimed in~\cite{richter2016gta} that experts spend up to 90 minutes per image for pixel-wise annotation.
Some weakly supervised approaches are proposed to leverage easy-attained annotations like image level label~\cite{huang2018dsrg,li2019weaklier}, bounding boxes~\cite{dai2015boxsup,khoreva2017simple} or scribbles~\cite{lin2016scribblesup} to circumvent the expensive cost of pixel-wise annotations.
Another direction addressing the annotation problem is to leverage synthetic datasets, {\textit e.g.}, GTA5~\cite{richter2016gta} and SYNTHIA~\cite{ros2016synthia}. Though synthetic data with annotations is rather easy to collect, the performance of models trained on synthetic data drops drastically when applied on real scene due to the domain shift between synthetic and realistic data.

\subsection{Pairwise affinity.}
\label{subsec:pa}
Pairwise pixel affinity has a long history use in computer vision tasks. In early vision~\cite{poggio1986earlyvision}, local affinity has been utilized to characterize the intrinsic geometric structure. Pairwise affinity has been involved in segmentation as clustering cues in~\cite{shi2000normalcut,wang2015globallocal}. Recently, pairwise pixel affinity has been combined with convolutional neural network (CNN) to provide structure reasoning for semantic or instance segmentation. Example works are~\cite{chen2018deeplab,zheng2015conditional,ke2018aaf,liu2017spn,liu2018affinityderivation}. In~\cite{chen2018deeplab,zheng2015conditional}, the authors leverage pairwise affinity via conditional random field (CRF). An adaptive loss built on affinity is proposed in~\cite{ke2018aaf} to improve segmentation accuracy. In~\cite{liu2017spn}, the authors propose spatial propagation networks to learn affinity for improving semantic segmentation. In~\cite{liu2018affinityderivation}, the authors explicitly regress pairwise affinity, serving to separate instances of the same semantic for instance segmentation.

\begin{figure*}[tb]
    \centering
   \includegraphics[width=0.9\linewidth]{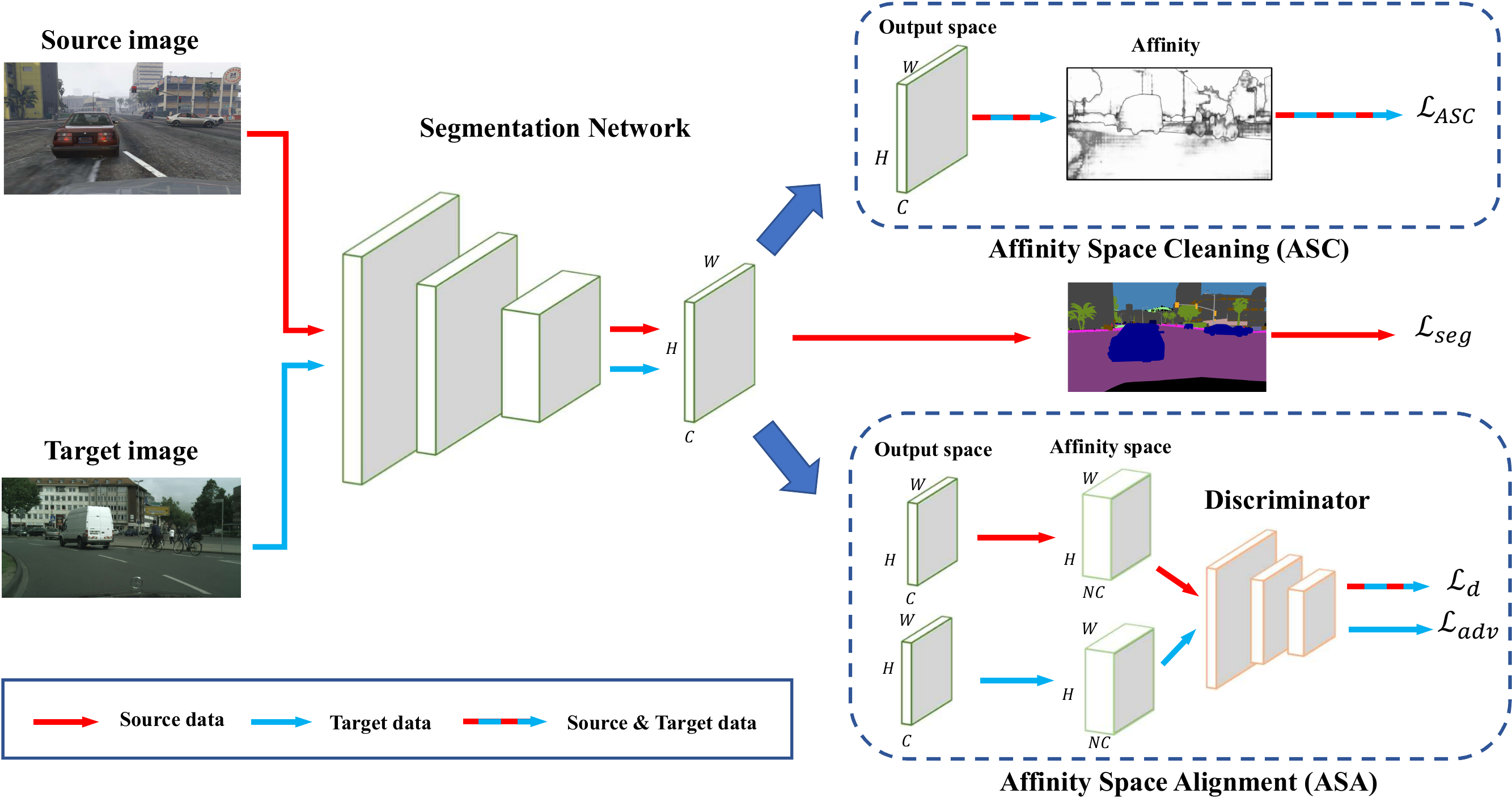}
    \caption{Overview of the proposed approach. For affinity space adaptation with affinity space cleaning, an affinity-based loss $\mathcal{L}_{ASC}$ is imposed on both source and target domain. For adversarial affinity space alignment strategy, we utilize adversarial training to make affinity space look similar across domains. \textcolor{red}{Red} arrows are used for source domain, \textcolor{blue}{blue} arrows are used for target domain and the mixed arrows represent both domains. For both affinity space adaptation schemes, a segmentation loss $\mathcal{L}_{seg}$ is also computed between segmentation predictions and provided annotations on source images. }
    \label{fig:arch}
\end{figure*}

\subsection{Domain adaptation}
\label{subsec:da}
Domain adaptation aims to address the domain-shift problem between the source and target domains.
Numerous methods~\cite{tzeng2017adda,zhang2018collaborative,saito2018mcd,xie2018learning,long2016unsupervised,saito2017adversarialdropout,li2018domain,li2019localitypreseve} are developed for image classification across domains.
The main idea is to minimize the difference of distributions across domains by adversarial learning.

Similar to adaptation on classification, most modern adaptation models for semantic segmentation also rely on adversarial learning to adapt domains either in intermediate feature level~\cite{hoffman2016fcns,hong2018cganl,wang2019weaklyda,luo2019siban} or output level~\cite{tsai2018adaptsegnet,vu2018advent,chen2019maxsquare}. UDA for semantic segmentation was firstly addressed in~\cite{hoffman2016fcns} that aligns the global features of source and target domains via adversarial network. Recently, in~\cite{luo2019siban}, a significance-aware information bottleneck is introduced to align domains in latent feature space. It has been shown in~\cite{tsai2018adaptsegnet,vu2018advent,chen2019maxsquare,tsai2019sodpr} that aligning domains in output level may be a better choice than feature level for semantic segmentation across domains. In~\cite{tsai2018adaptsegnet}, the authors first propose to explicitly align domains on the structured output, which is then improved or extended by introducing new loss functions~\cite{vu2018advent,chen2019maxsquare} or extra patch-level output space alignment~\cite{tsai2019sodpr}.

In addition to the distribution (\ie, feature or output level) alignment strategies that regard each semantic category equally, some approaches~\cite{Yawei2019clan,du2019ssfdan,tsai2019sodpr} tend to boost the performance for some specific categories or regions. In~\cite{Yawei2019clan}, Luo~\etal propose a category-level adversarial network which adaptively weights the adversarial loss for category-level adaptation. SSF-DAN~\cite{du2019ssfdan} utilizes pseudo labels to get semantic features of each class for separated adaptation, which brings performance gain on less frequent classes. In~\cite{tsai2019sodpr}, Tsai \textit{et al.} take multi modes of patch-wise output distribution into consideration and attach an additional patch alignment module to~\cite{tsai2018adaptsegnet}, further improving the adaptation performance. These recent methods suggest that class- or region-conditioned adaptation is beneficial for semantic segmentation across domains based on distribution alignment.

Techniques like image translation and self-training have been proved
useful or complementary to the distribution alignment based methods for domain adaptation in many works~\cite{hoffman2017cycada,wu2018dcan,zou2018cbst,chang2019ABS,li2019BDL,tsai2019sodpr}. In~\cite{hoffman2017cycada}, the authors utilize CycleGAN~\cite{CycleGAN2017} to generate extra target data in addition to adversarial adaptation on the feature level. DCAN~\cite{wu2018dcan} performs image-level and feature-level adaptation at the same time. DISE~\cite{chang2019ABS} disentangles images into domain-invariant structure and domain-specific textures for better image translation and label transfer. In CBST~\cite{zou2018cbst}, the authors propose a novel class-balanced self-training framework, where imbalanced class distribution and spatial priors are taken into consideration. More recently, BDL~\cite{li2019BDL} proposes a bidirectional learning framework to make the image translation model, the segmentation adaptation model, and self-training model learn alternatively, further boosting the performance.

Some other methods seek to reduce domain gap via effective utilization of source data~\cite{sadat2018effectiveuse}, designing loss function~\cite{zhu2018ptp,chen2019maxsquare}, curriculum model adaptation~\cite{sakaridis2018cmadal,sakaridis2018foggy}, or self-ensembling strategy~\cite{SEAN, choi2019segan}

\subsection{Comparison with related works}
\label{subsec:comparison}
Most existing approaches for unsupervised domain adaptation focus on aligning domains based on individual pixel-wise information in image, intermediate feature, or output level.
Though these approaches have achieved remarkable performance across domains, the structural and contextual representation is not explicitly exploited.
We propose to highlight structure through leveraging co-occurring patterns between pairwise pixels in the output level of structured semantic segmentation.
The proposed approach performs domain adaptation on the affinity space given by the affinity relationship between adjacent pixels in the output prediction.
We exploit two affinity space adaptation strategies: affinity space cleaning and adversarial affinity space alignment.
Though the proposed method is driven by the importance of pairwise affinity involved in some fully supervised segmentation methods, we leverage the pairwise affinity in a different way by constructing an affinity space and performing domain adaptation on it.
This demonstrates for the first time that the affinity relationship is beneficial for UDA in semantic segmentation.

\section{Approach}
\label{sec:approach}

\subsection{Overview}
\label{subsec:overview}

Given the source data $\mathbb{X}_S \subset \mathbb{R}^{H \times W \times 3}$ with dense pixel-wise annotations $\mathbb{Y}_S \subset (1, \dots, C)^{H \times W}$ and the target data $\mathbb{X}_T \subset \mathbb{R}^{H \times W \times 3}$ without annotations.
Unsupervised domain adaptation (UDA) in semantic segmentation aims to train a model that performs well on the target data $\mathbb{X}_T$, where $H$ and $W$ is the image height and width, respectively, and $C$ is the number of classes.
Considering that the semantic segmentation is a structured prediction problem, we leverage the pairwise co-occurring patterns that reveals the output structure being invariant across domains.
Specifically, we introduce the concept of affinity space which is built upon affinity between adjacent pixels in the output space.
We perform UDA for semantic segmentation on the affinity space instead of on information (from image, feature, or output level) of individual pixel.
To this end,  we propose two schemes to adapt based on the concept of affinity space,
which describes the similarity between adjacent pixels and has different implementations depending on the proposed schemes.
The first strategy implicitly regularizes the affinity space via an affinity space cleaning loss (see Section~\ref{subsec:asc}).
The second one is an adversarial framework that explicitly aligns affinity distribution across domains (see Section~\ref{subsec:asa}).
The pipeline of the proposed affinity space adaptation framework is illustrated in Fig.~\ref{fig:arch}.
The adopted network architecture for performing the semantic segmentation across domains is depicted in Section~\ref{subsec:networkarch}.

\subsection{Affinity space cleaning}
\label{subsec:asc}

The semantic segmentation can be regarded as a structured prediction problem. Even though there is no ground-truth for the target images, we may still impose some constraints on the output structures. Indeed, based on the observation (see Fig.~\ref{fig:introaff}) about affinity between adjacent pixels in the output space, we notice that the affinity is unclean in the sense that it is low on most part of target images without adaptation. For an expected semantic segmentation, the affinity is low only on edges of adjacent semantic regions, resulting in a clean affinity map. Therefore, we propose an affinity space cleaning loss on both source and target images to force the network to produce clean affinity map of output space. For an image $X$, the proposed affinity space cleaning (ASC) loss is formulated as following:

\begin{equation}
    \mathcal{L}_{ASC}(X) =  \frac{1}{|X|}\sum_{x \in X}\sum_{n \in \mathcal{N}(x)}{1 - \frac{P_x \cdot P_n}{{\left\|P_x\right\|}{\left\|P_n\right\|}
    }},
    \label{eq:ascloss}
\end{equation}
% \begin{equation}
%     \label{equ:crossentropy}
%     \mathcal{L}_{seg} = -\sum_{h, w}\sum_{c \in C}Y_{s}^{(h,w,c)}\log(P_s^{(h,w,c)})
% \end{equation}
where $ \mathcal{N}(x)$ is the set of 4 or 8 spatially adjacent neighbors of pixel $x$, $|\cdot|$ denotes the cardinality, $P$ stands for the Softmax output prediction, and $\|\cdot\|$ is the magnitude of the vector.
Minimizing such affinity space cleaning loss encourages high affinity everywhere.
Though such ASC on border pixels does not entirely make sense, in most cases, pairwise adjacent pixels lying on the same semantic region dominate over adjacent pixels from two different semantic regions in the expected segmentation. Incorporating the structure reasoning in this way is reasonable and regularizes the affinity space of the target domain to approach that of the source domain. For a target image $X_t$, we minimize such affinity cleaning loss. For a source image $X_s$ with ground-truth annotation $Y_s$, we also optimize the following cross-entropy loss for segmentation:
\begin{equation}
    \mathcal{L}_{seg}(X_s) = -\frac{1}{|X_s|}\sum_{h, w}\sum_{c \in C}Y_{s}^{(h,w,c)}\log(P_{s}^{(h,w,c)}),
    \label{eq:segloss}
\end{equation}
where $Y_s$ is the one-hot representation of ground-truth annotation. The final objective function to be minimized for this affinity space adaptation scheme is given by:
\begin{equation}
    \mathcal{L} = \mathcal{L}_{seg}{(X_s)} + \lambda_{ASC}\mathcal{L}_{ASC}{(X_s)} + \lambda_{ASC}\mathcal{L}_{ASC}{(X_t)},
     \label{eq:asctotalloss}
\end{equation}
where $\lambda_{ASC}$ is a weighting factor (set to a small value, \eg, 0.001) that represents the importance of loss $\mathcal{L}_{ASC}$.
The ASC loss in Eq.~\eqref{eq:ascloss} acts somehow as smoothing the region in target images without supervision, and is in the range of [0, 1]. In the beginning of training when the target prediction is not accurate, the cross-entropy loss for segmentation $\mathcal{L}_{seg}(X_s)$ on the source images is much larger than $\lambda_{ASC} = 0.001$. Thus, the beginning of training is mainly guided by the segmentation loss on the source images. When the segmentation loss on the source image drops to a relatively small value, the segmentation model already possesses a certain ability in segmenting the target images, and the ASC begins to influence the network training. Therefore, the proposed ASC strategy in general does not harm the model training.

Another interpretation of the proposed ASC may be that by optimizing on the source images in a fully supervised way, the segmentation model already possesses a certain ability in predicting semantic labels on target images. By further minimizing $\mathcal{L}_{ASC}$, we enforce the adjacent pixels to have the same label prediction, which can somehow propagate accurate classification score from current pixel to nearby pixels belonging to the same semantic region.

\subsection{Affinity space alignment}
\label{subsec:asa}

We also propose an adversarial framework to explicitly align the affinity space distribution of the target domain to that of the source domain. In this way, we could transfer knowledge of co-occurring patterns from the source domain to the target domain, \textit{e.g.} the sky class is always on the top of the building class and the rider class always comes up with bike or motorcycle class. For such adversarial affinity space alignment (ASA), we begin with constructing the affinity space $A \in \mathbb{R}^{H \times  W \times NC}$ from the output prediction $P \in \mathbb{R}^{H \times W \times C}$, where $N$ (4 or 8 for 4-connectivity or 8-connectivity) is the number of involved adjacent pixels on each pixel. Precisely, for a pixel $x$ and one of its neighboring pixel $n$, we denote $P_x = (p_x^1, \cdots, p_x^C)$ and $P_n = (p_n^1, \cdots, p_n^C)$ as the corresponding $C$-class Softmax output vector. We then build the affinity space $A$ based on affinity related vector $A_n = (a_n^1, \cdots, a_n^C)$ using KL divergence-like measure on each class, where
$a_n^i = (p_x^{i}\log(\frac{p_x^{i}}{p_n^{i}}) + {(1-p_x^{i})}\log(\frac{1-p_x^{i}} {1-p_n^{i}})$ for $i$-th class.
For $N$ pairs of neighboring pixels of underlying pixel $x$, we have $N$ such affinity vectors, resulting in a vector of $NC$ channels by concatenating them together $A = (A_1, \cdots, A_N)$. This gives rise to an affinity space $A$ of size $H \times W \times NC$ for each image $X$. Such KL divergence-like measure on each class preserves better the co-occurring information between different classes than K-way KL divergence and Cosine similarity which squeeze all channels into a scalar similarity between neighboring pixels.
For example, adjacent pixels of different K-way classification probability may result in the same value for these scalar measurements. With the proposed KL divergence-like measure on each class, different vectors are produced.

Given an image $X$ from the source domain or target domain, the segmentation network produces an affinity space $A$. We forward the affinity space $A$ to a fully-convolutional discriminator $\boldsymbol{D}$ using a binary cross-entropy loss $\mathcal{L}_{d}$ for the source and target domains. The loss $\mathcal{L}_d$ is formulated as:
\begin{equation}
\begin{aligned}
\mathcal{L}_{d}(A) = -\sum_{h, w} \big( (1-z)log(\boldsymbol{D}(A)^{(h, w, 0)}) \\ +zlog(\boldsymbol{D}(A)^{(h, w, 1)}) \big ),
\label{equ:dloss}
\end{aligned}
\end{equation}
where $z$ equals 0 for sample from target domain, and $z$ is 1 for sample from the source domain.
For an image $X_t$ from the target domain, the adversarial objective to train the segmentation network is given by:
\begin{equation}
    \mathcal{L}_{adv}({A_{t}}) = -\sum_{h, w} \log(\boldsymbol{D}(A_{t})^{(h, w, 1)}).
    \label{equ:adv_loss}
\end{equation}
The goal is to train the segmentation network and fool the discriminator by maximizing the probability of affinity space $A_t$ of the target domain being considered as the affinity space of the source domain. Collaborating with the segmentation loss in Eq.~\eqref{eq:segloss} on source images, the overall loss function to train the segmentation network can be written as:
\begin{equation}
    \mathcal{L}_{ASA}(X_s, X_t) = \mathcal{L}_{seg}(X_s) + \lambda_{ASA} \mathcal{L}_{adv}(A_t),
    \label{eq:asaloss}
\end{equation}
where $\lambda_{ASA}$ is the weighting factor for the adversarial term $\mathcal{L}_{adv}$. During training, we alternatively optimize discriminator network $\boldsymbol{D}$ and the segmentation network using loss function in Eq.~\eqref{equ:dloss} and Eq.~\eqref{eq:asaloss}, respectively.

\subsection{Network architecture}
\label{subsec:networkarch}
For the semantic segmentation network, we adopt the same network architecture as~\cite{tsai2018adaptsegnet}. More specifically, we adopt Deeplab-V2 model~\cite{chen2018deeplab} as the base architecture while discarding the multi-scale fusion strategy. We evaluate the proposed method with two CNN backbones: VGG-16~\cite{vgg16network} and ResNet-101~\cite{he2016resnet}, both of which are initialized with the model pre-trained on ImageNet~\cite{russakovsky2015imagenet}. Following~\cite{chen2018deeplab}, we modify the stride and dilation rate of the last two convolutional layers of CNN backbones. Atrous Spatial Pyramid Pooling (ASPP) with dilation rates \{6, 12, 18, 24 \} is also applied on the last layer of both backbone models. For the discriminator, we choose the similar network architecture used in~\cite{tsai2018adaptsegnet} for the proposed adversarial affinity space alignment. More precisely, the discriminator is composed of five $4\times4$ convolution layers with stride 2 and channel numbers \{64, 128, 256, 512, 1\}, respectively. Each convolutional layer, except the last one, is followed by a leaky-ReLU layer with a fixed negative slope of 0.2.

\begin{table*}[t]
  \caption{Quantitative comparison with some state-of-the-art methods on adapting GTA5 to Cityscapes. In the first and second groups, VGG-16 and ResNet-101 backbone networks are adopted, respectively. Methods trained using multi-level adaptation are marked with $\dag$.}
  \label{table:gta5_all}
  \centering
  \resizebox{\textwidth}{!}{
   \begin{tabular}{lcccccccccccccccccccc}
    \toprule
    \multicolumn{21}{c}{GTA5 $\rightarrow$ Cityscapes} \\
		\midrule
		Method & \rotatebox{90}{road} & \rotatebox{90}{sidewalk} & \rotatebox{90}{building} & \rotatebox{90}{wall} & \rotatebox{90}{fence} & \rotatebox{90}{pole} & \rotatebox{90}{light} & \rotatebox{90}{sign} & \rotatebox{90}{veg} & \rotatebox{90}{terrain} & \rotatebox{90}{sky} & \rotatebox{90}{person} & \rotatebox{90}{rider} & \rotatebox{90}{car} & \rotatebox{90}{truck} & \rotatebox{90}{bus} & \rotatebox{90}{train} & \rotatebox{90}{mbike} & \rotatebox{90}{bike} & mIoU\\
		\midrule
		FCNs in the Wild~\cite{hoffman2016fcns}&70.4&32.4&62.1&14.9&5.4&10.9&14.2&2.7&79.2&21.3&64.6&44.1&4.2&70.4&8.0&7.3&0.0&3.5&0.0&27.1\\
		CyCADA~\cite{hoffman2017cycada}&83.5&\textbf{38.3}&76.4&20.6&16.5&22.2&\textbf{26.2}&\textbf{21.9}&80.4&28.7&65.7&\textbf{49.4}&4.2&74.6&16.0&26.6& 2.0 &8.0&0.0&34.8\\
		AdaptSegNet~\cite{tsai2018adaptsegnet}& \textbf{87.3} &29.8&78.6&21.1&18.2&22.5&21.5&11.0&79.7&29.6&71.3&46.8&6.5&80.1&23.0&\textbf{26.9}&0.0&10.6&0.3&35.0 \\
		ADVENT~\cite{vu2018advent} & 86.9 & 28.7 & 78.7 &\textbf{28.5}&\textbf{25.2}&17.1&20.3&10.9&80.0&26.4&70.2&47.1&8.4&\textbf{81.5}& \textbf{26.0} & 17.2 & \textbf{18.9} & 11.7 & 1.6 & \textbf{36.1} \\
    SIBAN~\cite{luo2019siban} & 83.4 & 13.0 & 77.8 & 20.4 & 17.5 & \textbf{24.6} & 22.8 & 9.6 & \textbf{81.3} & 29.6 & \textbf{77.3} & 42.7 & 10.9 & 76.0 & 22.8 & 17.9 & 5.7 & 14.2 & 2.0 & 34.2 \\
    Source Only (our) & 54.5 & 15.9 & 64.5 & 13.6 & 19.1 & 16.9 & 20.0 & 6.5 & 77.3 & 9.4 & 74.9 & 46.3 & 10.6 & 47.4 & 11.2 & 13.2 & 0.1 & 11.6 & 0.8 & 27.1 \\
    ASC (our)& 85.2 & 7.8 & 75.5 & 14.0 & 22.6 & 17.8 & 22.8 & 10.0 & 78.2 & 21.1 & 59.9 & 48.3 & 9.2 & 79.9 & 21.2 & 23.4 & 0 & \textbf{18.0} & 0.7 & 32.4 \\
    ASA (our)& 86.9 & 32.5 & \textbf{79.0} & 22.8 & 23.1 & 20.7 & 22.0 & 12.6 & 80.0 & \textbf{32.2} & 68.5 & 43.6 & \textbf{11.9} & 81.3 & 20.8 & 9.6 & 4.2 & 16.9 & \textbf{8.5} & 35.6\\
		\midrule
		AdaptSegNet~\cite{tsai2018adaptsegnet} & 86.5 & 25.9 & 79.8 & 22.1 & 20.0 & 23.6 & 33.1 & 21.8 & 81.8 & 25.9 & 75.9 & 57.3 & 26.2 & 76.3 & 29.8 & 32.1 & 7.2 & 29.5 & 32.5 & 41.4\\
    AdaptSegNet$^\dag$~\cite{tsai2018adaptsegnet} & 86.5 & 36.0 & 79.9 & 23.4 & 23.3 & 23.9 & 35.2 & 14.8 & 83.4 & 33.3 & 75.6 & 58.5 & 27.6 & 73.7 & 32.5 & 35.4 & 3.9 & 30.1 & 28.1 & 42.4\\
	  ADVENT$^\dag$~\cite{vu2018advent} &\textbf{89.9}&\textbf{36.5}&\textbf{81.6}&\textbf{29.2}&\textbf{25.2}&28.5&32.3& \textbf{22.4} &\textbf{83.9}&34.0&\textbf{77.1}&57.4&27.9&83.7&29.4&39.1&1.5&28.4&23.3&43.8\\
    SIBAN~\cite{luo2019siban} & 88.5 & 35.4 & 79.5 & 26.3 & 24.3 & 28.5 & 32.5 & 18.3 & 81.2 & \textbf{40.0} & 76.5 & 58.1 & 25.8 & 82.6 & 30.3 & 34.4 & 3.4 & 21.6 & 21.5 & 42.6 \\
    MaxSquare~\cite{chen2019maxsquare} & 88.1 & 27.7 & 80.8 & 28.7 & 19.8 & 24.9 & 34.0 & 17.8 & 83.6 & 34.7 & 76.0 & 58.6 & 28.6 & 84.1 & 37.8 & 43.1 & 7.2 & 32.2 & 34.2 & 44.3 \\
    Source Only (our) & 84.2 & 9.4 & 74.6 & 14.6 & 17.9 & 22.4 & 27.8 & 17.8 & 69.1 & 12.4 & 73.0 & 55.3 & 27.0 & 82.6 & 30.9 & 37.9 & 3.2 & 27.6 & 35.6 & 38.1 \\
    ASC (our)& 85.5 & 23.8 & 79.8 & 18.9 & 21.1 & 27.5 & \textbf{36.6} & 18.8 & 82.7 & 25.9 & 76.0 & \textbf{59.7} & 29.0 & 81.7 &33.1 & 43.5 & 14.1 & 33.7 & \textbf{41.4}  & 43.8\\
    ASA (our)& 89.2 & 27.8 & 81.3 & 25.3 & 22.7 & \textbf{28.7} & 36.5 & 19.6 & 83.8 & 31.4 & \textbf{77.1} & 59.2 & \textbf{29.8} & \textbf{84.3} & \textbf{33.2} & \textbf{45.6} & \textbf{16.9}  & \textbf{34.5} & 30.8 & \textbf{45.1}\\
    \bottomrule
  \end{tabular}
  }
\end{table*}

\section{Experiments}
\label{sec:experiments}
To validate the effectiveness of the proposed approach for semantic segmentation across domains, we conduct experiments on two popular domain adaptation benchmarks: GTA5~\cite{richter2016gta} to Cityscapes~\cite{cordts2016cityscapes} and SYNTHIA~\cite{ros2016synthia} to Cityscapes. Experimental results, including cross-city case and medical image segmentation across domains, are also given for validation on real-to-real adaptation scenarios.
In the following, we clarify the implementation details and compare the proposed method with other state-of-the-art approaches using similar backbones.

\subsection{Datasets \& evaluation metrics}
\label{subsec:dataset}
\noindent\textbf{Synthetic-to-Real adaptation.}
We first conduct experiments on synthetic-to-real adaptation using Cityscapes~\cite{cordts2016cityscapes} as the target dataset, GTA5~\cite{richter2016gta} and SYNTHIA~\cite{ros2016synthia} as the source dataset, respectively. The details about these datasets are shortly described as follows.

\smallskip
\textit{Cityscapes~\cite{cordts2016cityscapes}} is a widely used dataset for semantic segmentation, which contains high-quality dense annotations of 5000 images collected from 50 cities. The dataset is divided into 2975, 500, and 1525 images for training, validation, and testing, respectively. It provides 30 common classes, of which 19 classes are used for evaluation.

\smallskip
\textit{GTA5~\cite{richter2016gta}} is a synthetic dataset containing 24966 images built with the gaming engine Grand Theft Auto V. Pixel-level semantic annotations of 33 classes are provided.
Following other methods, we only use the common 19 classes to pair with the  Cityscapes~\cite{cordts2016cityscapes} dataset.

\smallskip
\textit{SYNTHIA~\cite{ros2016synthia}} is another large synthetic dataset rendered by the Unity game engine.
Its subset, named SYNTHIA-RAND-CITYSCAPES, provides 9400 images with labels compatible with the Cityscapes~\cite{cordts2016cityscapes} classes.
Different from GTA5, this dataset contains synthetic images with various viewpoints, making it challenging for domain adaptation.

Following previous works~\cite{zou2018cbst,tsai2018adaptsegnet,vu2018advent}, we utilize labeled synthetic images and unlabeled training images of Cityscapes to perform adaptation and evaluate the proposed adapted model on Cityscapes `val' split.

\medskip
\noindent\textbf{Real-to-Real adaptation.}
We also evaluate on real-to-real adaptation scenarios to further validate the effectiveness of the proposed method. For that, we conduct experiments on cross-city dataset~\cite{chen2017nmd} and two retinal fundus image datasets: REFUGE~\cite{orlando2020refuge} and RIM-ONE-r3~\cite{fumero2011rim}. The details of these datasets are shortly given in the following.

\smallskip
\textit{Cross-city~\cite{chen2017nmd}} is a high-quality road scene dataset collected from four different cities: Rome, Rio, Tokyo, and Taipei.
Each city consists of 3200 unlabeled images and 100 Cityscapes-compatible annotated images.
Similar to~\cite{chen2017nmd}, we adopt the Cityscapes training set as the source domain and adapt the segmentation model to each target city using 3200 unlabelled images. The other 100 annotated images of each dataset are used for evaluation.

\smallskip
\textit{REFUGE~\cite{orlando2020refuge} and RIM-ONE-r3~\cite{fumero2011rim}} are two retinal fundus image datasets for retinal optic disk and cup segmentation.
REFUGE~\cite{orlando2020refuge} has 400 high resolution training images. RIM-ONE-R3~\cite{fumero2011rim} dataset contains 99 training samples and 60 samples for testing. Both datasets have precise pixel-wise segmentation annotations for optic disk and cup. We use the 400 high resolution training images in REFUGE~\cite{orlando2020refuge} as the source domain, and adapt the segmentation model to 99 samples of the RIM-ONE-R3~\cite{fumero2011rim} dataset. We report segmentation results on the 60 test images in RIM-ONE-R3~\cite{fumero2011rim} dataset.

\medskip
\noindent\textbf{Evaluation metric.} In all experiments, we adopt the  widely used Intersection-over-Union (IoU) for both synthetic-to-real and real-to-real segmentation adaptation on road scenes and Dice coefficients (DSC) for segmentation adaptation on medical images. Concretely, the two evaluation metrics are given by:
\begin{eqnarray}
    %\label{equ:iou}
    IoU \, & = &\, \frac{TP}{TP+FP+FN}, \\
%\end{equation}
%\begin{equation}
    DSC \, & = & \, \frac{2TP}{2TP+FP+FN},
    \label{equ:ioudsc}
\end{eqnarray}
where $TP$, $FP$, and $FN$ represent the number of true positive, false positive, and false negative pixels, respectively.

\subsection{Implementation details}
\label{subsec:implementation}
The implementation is based on the public toolbox PyTorch~\cite{paszke2017pytorch}. All experiments are performed on a workstation with an NVIDIA TITAN Xp GPU card of 12GB memory. For a fair comparison, we use the same hyper-parameter settings with AdaptSegNet~\cite{tsai2018adaptsegnet}. More specifically, for the segmentation model, we use SGD as the optimizer with ``poly'' learning rate policy where the learning rate equals to $base\_lr*{(1-\frac{iter}{total\_iter})}^{power}$. We set the initial learning rate $base\_lr$ to $2.5\times10^{-4}$ and the $power$ to 0.9. The momentum is set to 0.9 and the weight decay is set to 0.0005. For the discriminator, we employ Adam optimizer with an initial learning rate  $1\times{10}^{-4}$. The momentum is set as 0.9 and 0.99. The hyper-parameter ${\lambda}_{ASC}$ in Eq.~\eqref{eq:asctotalloss}, and ${\lambda}_{ASA}$ in Eq.~\eqref{eq:asaloss} are all set to $0.001$ for all experiments in this paper. Except explicitly stated, we use 8-connectivity adjacent pixels to define affinity space. Due to limited GPU memory, during training, the GTA5 (\textit{resp}. SYNTHIA) source images are resized to $720\times 1280$ (\textit{resp}. $ 760 \times 1280$), and the target Cityscapes images are resized to $512 \times 1024$ for all experiments. In the testing phase, the same size $512 \times 1024$ as the training phase for Cityscapes is used. The segmentation result is then resized to the original input size at evaluation time.
In the real-to-real adaptation experiments, we resize the images into  $512 \times 1024$ as the training input for cross-city adaptation, and the images are resized to $512 \times 512$ on both REFUGE and RIM-ONE-r3 dataset for segmentation adaptation on medical images.
Note that we do not use the ground truth to select the best adapted model. Instead, as we hypothesize in Section~\ref{subsec:asc}, an expected semantic segmentation network produces clean affinity of output space. Thus, we select the the adapted model by calculating the average pixel affinity on the train set (ground-truth-free). Specifically, we select the model which has the largest average pixel affinity value over the train set of the target domain.

\begin{figure*}[!t]
  \centering
  \includegraphics[width=1.0\linewidth]{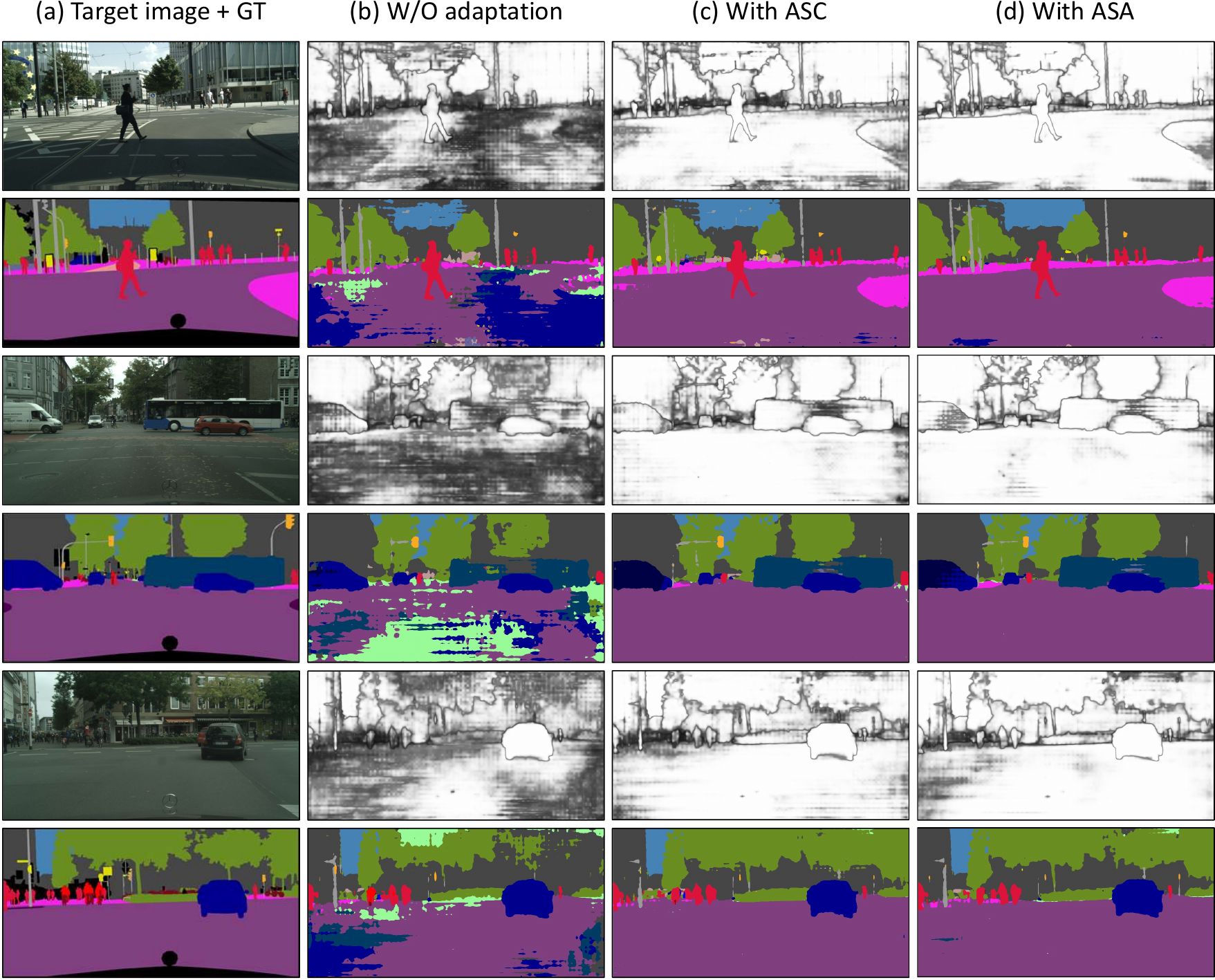}
  \caption{Qualitative results on GTA5 $\rightarrow$ Cityscapes adaptation. (a) Target images and the corresponding GTs. (b-d): affinity map and semantic segmentation results before adaptation in (b), using proposed affinity space cleaning in (c), and with proposed adversarial affinity space alignment in (d).}
  \label{fig:resvis}
\end{figure*}

\begin{table*}[t]
  \caption{Quantitative comparison with some state-of-the-art methods on adapting SYNTHIA to Cityscapes. In the first and second groups, VGG-16 and ResNet-101 backbone networks are adopted, respectively. Methods trained using multi-level adaptation are marked with $\dag$.}
  \label{table:synthia_all}
  \centering
  \resizebox{\textwidth}{!}{
  \begin{tabular}{lcccccccccccccccccc}
    \toprule
    \multicolumn{19}{c}{SYNTHIA $\rightarrow$ Cityscapes} \\
		\midrule
		Method & \rotatebox{90}{road} & \rotatebox{90}{sidewalk} & \rotatebox{90}{building} & \rotatebox{90}{wall} & \rotatebox{90}{fence} & \rotatebox{90}{pole} & \rotatebox{90}{light} & \rotatebox{90}{sign} & \rotatebox{90}{veg} & \rotatebox{90}{sky} & \rotatebox{90}{person} & \rotatebox{90}{rider} & \rotatebox{90}{car} & \rotatebox{90}{bus} & \rotatebox{90}{mbike} & \rotatebox{90}{bike} & mIoU & mIoU$^*$\\
		\midrule
		FCNs in the wild~\cite{hoffman2016fcns} & 11.5 & 19.6 & 30.8 & 4.4 & 0.0 & 20.3 & 0.1 & 11.7 & 42.3 & 68.7 & \textbf{51.2} & 3.8 & 54.0 & 3.2 & 0.2 & 0.6 & 20.2 & 22.1 \\
		AdaptSegNet~\cite{tsai2018adaptsegnet} & \textbf{78.9} & 29.2 & 75.5 & - & - & - & 0.1 & 4.8 & 72.6 & 76.7 & 43.4 & 8.8 & 71.1 & 16.0 & 3.6 & 8.4 &- & 37.6 \\
		ADVENT~\cite{vu2018advent} & 67.9 & \textbf{29.4} & 71.9 & 6.3 & 0.3 & 19.9  & 0.6 & 2.6 & 74.9 & 74.9 & 35.4 & 9.6 & 67.8 & 21.4 & 4.1 & \textbf{15.5} & 31.4 & 36.6 \\

    SIBAN~\cite{luo2019siban} & 70.1 & 25.7 & \textbf{80.9} & - & - & - & 3.8 & 7.2 & 72.3 & \textbf{80.5} & 43.3 & 5.0 & \textbf{73.3} & 16.0 & 1.7 & 3.6 & - & 37.2 \\

    Source Only (our) & 3.13 & 11.9 & 63.7 & 7.0 & 0.2 & 21.1 & 3.4 & 10.3 & 71.0  & 75.6 & 46.6 & 4.4 & 57.5  & 18.8 & 2.1 & 4.4 & 25.1 & 28.7\\
    ASC (our)& 6.1 & 11.9 & 76.0 & 2.4 & 0.2 & 20.6  & \textbf{7.5} & \textbf{13.6} & \textbf{76.9} & 75.7 & 50.7 & 12.5 & 70.0 & \textbf{21.5} & 3.1 & 8.0 & 28.5 & 33.3 \\
		ASA (our)& 72.6 & 24.2 & 74.2 & \textbf{8.6} & \textbf{0.6} & \textbf{21.3} & 6.1 & 12.6 & 73.7 & 77.0 & 42.3 & \textbf{13.0} & 67.9 & 19.1 & \textbf{6.0} & 14.3 & \textbf{33.3} & \textbf{38.7} \\
		\midrule
    AdaptSegNet~\cite{tsai2018adaptsegnet}& 79.2 & 37.2 & 78.8 & - & - & -  & 9.9 & 10.5 & 78.2 & 80.5 & 53.5 & 19.6 & 67.0 & 29.5 & \textbf{21.6} & 31.3 & - & 45.9 \\
		AdaptSegNet$^\dag$~\cite{tsai2018adaptsegnet}& 84.3 & 42.7 & 77.5 & 9.3 & 0.2 & 22.9 & 4.7 & 7.0 & 77.9 & 82.5 & 54.3 & 21.0 & 72.3 & 32.2 & 18.9 & 32.3 & 40.0 & 46.7  \\
		ADVENT$^\dag$~\cite{vu2018advent} & 87.0 & 44.1 & 79.7 & \textbf{9.6} & \textbf{0.6} & 24.3 & 4.8 & 7.2 & 80.1 & \textbf{83.6} & 56.4 & \textbf{23.7} & 72.7 & 32.6 & 12.8 & \textbf{33.7} & 40.8 & 47.6\\
    SIBAN~\cite{luo2019siban} & 82.5 & 24.0 & 79.4 & - & - & -  & \textbf{16.5} & \textbf{12.7} & 79.2 & 82.8 & 58.3 & 18.0 & 79.3 & 25.3 & 17.6 & 25.9 & - & 46.3 \\
    MaxSquare~\cite{chen2019maxsquare} & 77.4 & 34.0 & 78.7 & 5.6 & 0.2 & \textbf{27.7} & 5.8 & 9.8 & \textbf{80.7} & 83.2 & \textbf{58.5} & 20.5 & 74.1 & 32.1 & 11.0 & 29.9 & 39.3 & 45.8 \\
    Source Only (our) & 54.8 & 23.1 & 73.7 & 8.5 & 0.1 & 25.4 & 10.5 & 10.3 & 74.7 & 80.1 & 54.8 & 16.3 & 42.3 & 27.7 & 20.9 & 19.1 & 33.9 & 39.1\\
    ASC (our)& 77.6 & 31.2 & 75.1 & 4.3 & 0.2 & 26.5 & 9.7 & 10.9 & 78.9 & 81.6 & 55.2 & 19.6 & 76.4 & 26.1 & 16.9 & 29.3 & 38.7 & 45.3\\
    ASA (our)& \textbf{91.2} & \textbf{48.5} & \textbf{80.4} & 3.7 & 0.3 & 21.7 & 5.5 & 5.2 & 79.5 & \textbf{83.6} & 56.4 & 21.0 & \textbf{80.3} & \textbf{36.2} & 20.0 & 32.9 & \textbf{41.7}& \textbf{49.3}\\
    \bottomrule
  \end{tabular}
 }
\end{table*}

\subsection{Synthetic-to-real adaptation results}
The proposed method aims to perform adaptation by aligning affinity space of output prediction and thus falls into the category of distribution alignment (DA) based methods. Therefore, we first compare the proposed approach with some state-of-the-art methods attempting to leverage adversarial training to align feature or output distributions. We then explore the complementary of the proposed approach with image translation (IT) and self-training (ST) using pseudo labels, and compare with some related approaches which also make use of these techniques. Note that The related methods that use a different backbone network other than VGG16 or ResNet101 are not compared in this paper.

\medskip
\noindent\textbf{Comparison with related DA-based approaches.}
We first compare the proposed method with some related distribution alignment (DA) based approaches. The methods relying on class- or region-conditioned adaptation to boost the performance for some specific categories or regions, are not included in the comparison.

\smallskip
\textit{GTA5 $\rightarrow$ Cityscapes.} We first benchmark the proposed method on adapting GTA5 to Cityscapes. Tab.~\ref{table:gta5_all} depicts the performance comparison with several state-of-the-art methods.
The proposed affinity space adaptation achieves competitive or superior performance. Specifically, the proposed affinity space cleaning performs competitively with other methods using both VGG16 and ResNet101 backbones. Using ResNet101 backbone, the proposed ASC yields 43.8\% mIoU, outperforming multi-level setting of~\cite{tsai2018adaptsegnet} by 1.4\% and similar single-level setting of~\cite{tsai2018adaptsegnet} by 2.4\% in terms of mIoU. It is also noteworthy that the proposed ASC with the designed affinity loss does not rely on additional adversarial discriminator, leading to more efficient training. The proposed adversarial affinity space alignment further boosts the performance for both VGG16 and ResNet101 backbones. More precisely, the proposed ASA using ResNet101 backbone outperforms recent works which directly aligns the Softmax output~\cite{tsai2018adaptsegnet} or output entropy~\cite{vu2018advent} between the source domain and the target domain (under multi-level adaptation) by 2.7\% and 1.3\% mIoU, respectively. Compared with SIBAN~\cite{luo2019siban} and MaxSquare~\cite{chen2019maxsquare}, the proposed ASA records an improvement of 2.5\% mIoU and 0.8\% mIoU, respectively.
Some qualitative results of the proposed affinity space adaptation are given in Fig.~\ref{fig:resvis}. The proposed method effectively aligns the affinity space, resulting in adapted semantic segmentation across domains.

\smallskip
\textit{SYNTHIA $\rightarrow$ Cityscapes.}
We then benchmark the proposed method on adapting from SYNTHIA dataset to Cityscapes. Following prior works~\cite{tsai2018adaptsegnet, vu2018advent}, we report both 16 and 13-categories (marked with *) results in Tab.~\ref{table:synthia_all}. The same observation also holds as on the GTA5 to Cityscapes adaptation. The proposed ASC achieves results on par with several state-of-the-art methods.
Consistent to the results of adaptation from GTA5 to Cityscapes, the proposed affinity space adaptation with ASA strategy performs competitively with the state-of-the-art methods. Precisely, the proposed ASA using ResNet101 backbone improves multi-level setting of~\cite{tsai2018adaptsegnet} and~\cite{vu2018advent} by 2.6\% and 1.7\% mIoU, respectively, and outperforms the similar single-level setting of~\cite{tsai2018adaptsegnet} by 3.4\% mIoU. Compared with the method in~\cite{luo2019siban} and in~\cite{chen2019maxsquare}, the proposed ASA brings 3.0\% mIoU and 3.5\% mIoU improvement, respectively.
For VGG16 backbone, the proposed ASA achieves 1.1\% mIoU improvement against~\cite{tsai2018adaptsegnet}. The improvement of the proposed affinity space adaptation is less significant with VGG16 backbone than with ResNet101 backbone. This is probably because that the semantic segmentation using VGG16 is not as accurate as using ResNet101 on the source domain, leading to affinity space to be aligned far from the ground-truth affinity space. This makes the adaptation more challenging.

\medskip

It is noteworthy that most methods do not perform well on the under-represented classes (\eg, fence, pole and light) shown in Tab.~\ref{table:gta5_all} and Tab.~\ref{table:synthia_all}. This is a common class imbalance issue in unsupervised domain adaptation. During adaptation, the network tends to pay more attention to the dominated classes than other classes. Consequently, the proposed method brings consistent performance improvements on the high frequent classes (e.g., road, building and sky) but not on the infrequent classes (e.g., fence, pole and light). Despite this, as depicted in Tab.~\ref{table:gta5_all} and Tab.~\ref{table:synthia_all}, the overall performance is still very encouraging. It is likely that combining with class balanced techniques may achieve better performances, which is a promising direction to be explored in the future.

\begin{table}
    \centering
    \caption{Complementary of the proposed ASA relying on distribution alignment (DA) with image translation (IT) and self-training (ST).
    }
    \small
    \begin{tabular}{l|ccc|c}
         \multicolumn{5}{c}{(a) GTA5 $\rightarrow$ Cityscapes} \\
         \bottomrule
         Method & IT  & DA  & ST  & mIOU  \\
         \hline
         DCAN ~\cite{wu2018dcan} & \checkmark & \checkmark &  & 41.7\\
         CBST~\cite{zou2018cbst}& & & \checkmark&41.5 \\
         DISE~\cite{chang2019ABS} & \checkmark & \checkmark &  & 45.4\\
         SODPR ~\cite{tsai2019sodpr} & \checkmark & \checkmark & \checkmark & 46.5\\
         BDL~\cite{li2019BDL} & \checkmark & \checkmark & \checkmark  & 48.5 \\
         \hline
         ASA (our)& & \checkmark & &45.1 \\
         ASA + IT (our) & \checkmark & \checkmark & & 45.6 \\
         ASA + ST (our)& & \checkmark & \checkmark&48.1 \\
         ASA + IT + ST (our)& \checkmark & \checkmark & \checkmark& \textbf{49.1} \\
         \toprule
    \end{tabular}

    \begin{tabular}{l|ccc|c}
         \multicolumn{5}{c}{(b) SYNTHIA $\rightarrow$ Cityscapes} \\
         \bottomrule
         Method& IT & DA & ST & mIOU \\
         \hline
         DCAN ~\cite{wu2018dcan} & \checkmark & \checkmark &  & 44.9\\
         CBST~\cite{zou2018cbst}& & & \checkmark&36.9 \\
         DISE~\cite{chang2019ABS} & \checkmark & \checkmark &  & 48.8\\
         SODPR ~\cite{tsai2019sodpr} & \checkmark & \checkmark & \checkmark & 46.5\\
         BDL~\cite{li2019BDL} & \checkmark & \checkmark & \checkmark  & 51.4 \\
         \hline
         ASA (our)& & \checkmark & &49.3 \\
         ASA + IT (our)  & \checkmark & \checkmark & & 49.8 \\
         ASA + ST (our)& & \checkmark & \checkmark&52.4 \\
         ASA + IT + ST (our)& \checkmark & \checkmark & \checkmark & \textbf{53.8} \\
         \toprule
    \end{tabular}
    \label{table:sota_rst}

\end{table}

\medskip
\noindent\textbf{Complementary with IT and ST.}
\label{sec:comp_sota}
Most recent works aim to solve the UDA semantic segmentation problem by leveraging the following three strategies: 1) image translation (IT) that bridges up the image style differences; 2) distribution alignment (DA) which aims to align distribution of intermediate feature or output prediction; 3) self-training that effectively uses pseudo labels. As demonstrated in~\cite{chang2019ABS,li2019BDL}, these three strategies are complementary to each other. Though the proposed method mainly focuses on distribution alignment (DA) of output prediction, we also discuss the complementary of the proposed ASA with IT and ST technique.

\smallskip
\textit{Image translation} is an intuitive way to reduce the domain gap on image level in unsupervised domain adaptation.
We first explore the complementary with IT by utilizing CyCADA~\cite{hoffman2017cycada} to generate target-like images from source domain images.
Then, we perform adaptation with ASA from new generated images to target images.
Since domian gap has been somehow reduced by image translation, the weight of ASA adaptation $\lambda_{ASA}$ in Eq.~\eqref{eq:asaloss} is set to 0.0001 in this experiment.
As depicted in Tab.~\ref{table:sota_rst}, the image translation technique brings performance gain to the proposed approach on both GTA5 $\rightarrow$ Cityscapes and SYNTHIA $\rightarrow$ Cityscapes adaptation, revealing that image translation is complementary to the proposed ASA adaptation.

\begin{table}
	\caption{
		Sensitivity analysis of loss weight $\lambda$ for ASC and ASA.
	}
	\label{tab:gta5_para}
	\small
	\centering
	\begin{tabular}{lccccc}
		\toprule
		& \multicolumn{4}{c}{GTA5 $\rightarrow$ Cityscapes} \\
		\midrule
		$\lambda$ & 0.0002 & 0.0005 & 0.001 & 0.002 & 0.004 \\
		\midrule
        $\lambda_{ASC}$ & 43.1 & 43.5 & \textbf{43.8} & 43.6 & 43.7 \\
		$\lambda_{ASA}$ & 44.9 & 45.0 & \textbf{45.1} & 44.7 & 44.6 \\
		\bottomrule
	\end{tabular}
\end{table}

\begin{figure}[!t]
  \centering
  \includegraphics[width=1.0\linewidth]{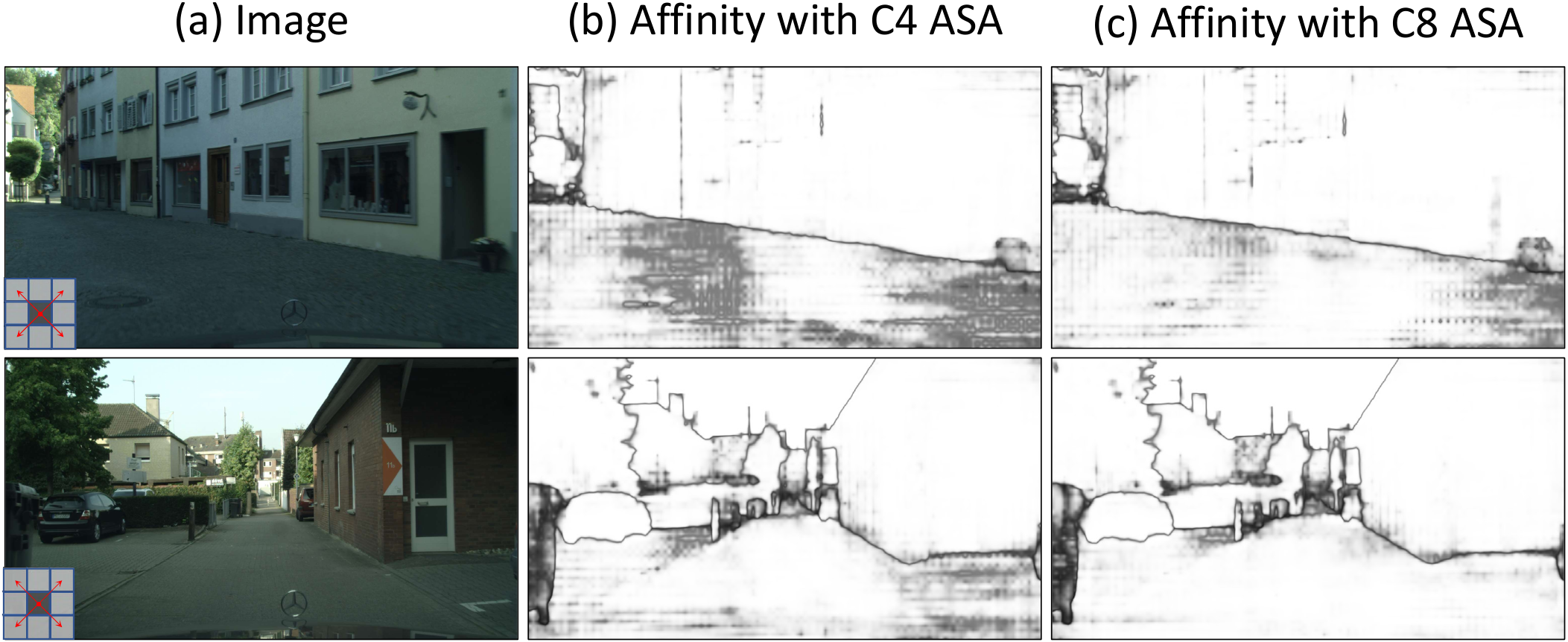}
  \caption{Diagonal affinity visualization for ASA with different connectivity settings on GTA5 $\rightarrow$ Cityscapes adaptation on some target images in (a). (b) and (c) show the diagonal affinity given by the average Cosine similarity between predictions of each pixel and all its diagonal neighboring pixels for ASA under 4-connectivity (C4) and 8-connectivity (C8) setting, respectively.}
  \label{fig:N_ablation}
\end{figure}

\begin{table}[t]
  \caption{Ablation study on the number of adjacent pixels $N$ involved in the proposed concept of affinity space and model complexity analysis on GTA5 $\rightarrow$ Cityscapes adaptation.}
  \label{table:ablation}
  \centering
  \resizebox{0.5\textwidth}{!}{
  \begin{tabular}{lcccc}
    \toprule
    Method & Connectivity ($N$) & GFLOPs & Memory(M) & mIoU \\
    \midrule
    Baseline & - & 644.9 & 9340 & 38.1\\
    AdaptSegNet~\cite{tsai2018adaptsegnet} & - & 1087.8 & 9602 & 41.4\\
    \midrule
    \multirow{2}{*}{ASC}
    &4 & 1041.7 & 9528 &42.6\\
    &8 & 1042.3 & 9532 & \textbf{43.8} \\
    \midrule
    \multirow{2}{*}{ASA}&4& 1129.3 & 9628 & 43.6 \\
    &8& 1167.7 & 9654 & \textbf{45.1} \\
    \bottomrule
  \end{tabular}
  }
\end{table}
\smallskip
\textit{Self-training} is another common strategy in unsupervised domain adaptation.
The core of self-training is how to get ``correct'' pseudo labels.
We take the model adapted with the proposed ASA to generate pseudo labels in a straightforward way.
Specifically, for each pixel $x$ in an image, we verify the Softmax output on that pixel.
If the highest confidence is larger than 0.9, we consider the predicted class as the pseudo label on $x$.
Otherwise, we simply set it to $ignored$ label, which will be ignored in gradient back-propagation.
We re-train the segmentation network with such pseudo labels on target domain.
As shown in Tab.~\ref{table:sota_rst}, the proposed approach is also complementary to the self-training strategy.
More specifically, on GTA5 $\rightarrow$ Cityscapes adaptation, the proposed ASA combined with self-training strategy yields 48.1\% mIoU, performing similarly with BDL~\cite{li2019BDL} which relies on extra image translation strategy. On SYNTHIA $\rightarrow$ Cityscapes adaptation, the proposed ASA combined with self-training strategy achieves 52.4\% mIoU, improving BDL~\cite{li2019BDL} by 1.0\% mIoU.

\begin{table*}
  \caption{Quantitative comparison with some state-of-the-art methods on adapting Cityscapes to Cross-City dataset.}
  \label{table:cross_city}
  \centering
  \resizebox{\textwidth}{!}{
		\begin{tabular}{l|l|ccccccccccccc|c}
			\bottomrule
		\multicolumn{16}{c}{Cityscapes $\rightarrow$ Cross-City} \\
			\hline
			City & Method & \rotatebox{90}{road} & \rotatebox{90}{sidewalk} & \rotatebox{90}{building} & \rotatebox{90}{light} & \rotatebox{90}{sign} & \rotatebox{90}{veg} & \rotatebox{90}{sky} & \rotatebox{90}{person} & \rotatebox{90}{rider} & \rotatebox{90}{car} & \rotatebox{90}{bus} & \rotatebox{90}{mbike} & \rotatebox{90}{bike} & mIoU\\
			\hline
    \multirow{4}{*}{Rome} &Cross-City \cite{chen2017nmd} & 79.5 & 29.3 & 84.5 & 0.0 & 22.2 & 80.6 & 82.8 & 29.5 & 13.0 & 71.7 & 37.5 & 25.9 & 1.0 & 42.9 \\
    & CBST~\cite{zou2018cbst} & \textbf{87.1} & \textbf{43.9} & \textbf{89.7} & 14.8 & \textbf{47.7} & 85.4 & 90.3 & 45.4 & \textbf{26.6} & \textbf{85.4} & 20.5 & 49.8 & \textbf{10.3} & 53.6 \\
	& AdaptSegNet~\cite{tsai2018adaptsegnet} & 83.9 & 34.2 & 88.3 & 18.8 & 40.2 & \textbf{86.2} & \textbf{93.1} & \textbf{47.8} & 21.7 & 80.9 & 47.8 & 48.3 & 8.6 & 53.8 \\
	&MaxSquare~\cite{chen2019maxsquare} & 80.0 & 27.6 & 87.0 & \textbf{20.8} & 42.5 & 85.1 & 92.4 & 46.7 & 22.9 & 82.1 & \textbf{53.5 }& \textbf{50.8} & 8.8 & 53.9\\
    % \cline{2-16}
    & Source Only (our) & 83.8 & 37.1 & 87.1 & 15.7 & 39.6 & 84.4 & 92.2 & 30.5 & 20.5 & 80.3 & 47.4 & 38.5 & 2.8 & 50.8  \\
	& ASA (our) & 84.9 & 35.1 & 87.5 & 17.3 & 40.3 & 85.3& 92.3 & 47.0 & 25.2 & 81.7 & 49.7 & 49.3 & 8.7 & \textbf{54.2} \\
    \hline
    \multirow{4}{*}{Rio} & Cross-City \cite{chen2017nmd} & 74.2 & 43.9 & 79.0 & 2.4 & 7.5 & 77.8 & 69.5 & 39.3 & 10.3 & 67.9 & 41.2 & 27.9 & 10.9 & 42.5 \\
    & CBST~\cite{zou2018cbst} & \textbf{84.3} & \textbf{55.2} & 85.4 & \textbf{19.6} & \textbf{30.1} & 80.5 & 77.9 & 55.2 & 28.6 & \textbf{79.7} & 33.2 & 37.6 & 11.5 & 52.2 \\
	& AdaptSegNet~\cite{tsai2018adaptsegnet} & 76.2 & 44.7 & 84.6 & 9.3 & 25.5 & \textbf{81.8} & 87.3 & \textbf{55.3} & 32.7 & 74.3 & 28.9 & 43.0 & 27.6 & 51.6 \\
	&MaxSquare~\cite{chen2019maxsquare} & 70.9 & 39.2 & \textbf{85.6} & 14.5 & 19.7 & \textbf{81.8} & \textbf{88.1} & 55.2 & 31.5 & 77.2 & 39.3 & 43.1 & 30.1 & 52.0 \\
% 	\cline{2-16}
    & Source Only (our) & 76.4 & 50.3 & 81.8 & 14.8 & 16.8 & 77.5 & 86.1 & 36.1 & 17.6 & 76.9 & \textbf{45.1} & 28.3 & 12.7 & 47.7  \\
	& ASA (our) & 79.1 & 49.8 & 84.9 & 14.3 & 20.8 & 80.2 & 87.6 & 54.9 & \textbf{33.7} & 78.3 & 40.0 & \textbf{50.9} & \textbf{32.0} & \textbf{54.4} \\
    \hline

    \multirow{4}{*}{Tokyo}
      & Cross-City \cite{chen2017nmd} & 83.4 & \textbf{35.4} & 72.8 & 12.3 & 12.7 & 77.4 & 64.3 & 42.7 & 21.5 & 64.1 & \textbf{20.8} & 8.9 & 40.3 & 42.8 \\
      & CBST~\cite{zou2018cbst} & \textbf{85.2} & 33.6 & \textbf{80.4} & 8.3 & \textbf{31.1} & \textbf{83.9} & 78.2 & 53.2 & 28.9 & \textbf{72.7} & 4.4 & 27.0 & 47.0 & 48.8 \\
   	  & AdaptSegNet~\cite{tsai2018adaptsegnet} & 81.5 & 26.0 & 77.8 & \textbf{17.8} & 26.8 & 82.7 & \textbf{90.9} & 55.8 & \textbf{38.0} & 72.1 & 4.2 & 24.5 & \textbf{50.8} & 49.9 \\
   	  & MaxSquare~\cite{chen2019maxsquare} & 79.3 & 28.5 & 78.3 & 14.5 & 27.9 & 82.8 & 89.6 & \textbf{57.3} & 31.9 & 71.9 & 6.0 & 29.1 & 49.2 & 49.7 \\
% 	  \cline{2-16}
      & Source Only (our) & 82.8 & 34.4 & 76.0 & 15.9 & 24.7 & 79.2 & 87.3 & 41.4 & 31.7 & 70.9 & 5.1 & 87.3 & 24.7 & 46.8  \\
	  & ASA (our) & 83.2 & 29.1 & 77.8 & 16.6 & 27.2 & 82.6 & 89.5 & 56.2 & 33.9 & \textbf{72.7} & 6.0 & 30.4 & 49.8 & \textbf{50.4} \\
      \hline
    \multirow{4}{*}{Taipei} & Cross-City \cite{chen2017nmd} & 78.6 & 28.6 & 80.0 & 13.1 & 7.6 & 68.2 & 82.1 & 16.8 & 9.4 & 60.4 & 34.0 & 26.5 & 9.9 & 39.6 \\
    & CBST~\cite{zou2018cbst} & \textbf{86.1} & \textbf{35.2} & 84.2 & 15.0 & \textbf{22.2} & 75.6 & 74.9 & 22.7 & \textbf{33.1} & \textbf{78.0} & 37.6 & \textbf{58.0} & 30.9 & \textbf{50.3} \\
    & AdaptSegNet~\cite{tsai2018adaptsegnet} & 81.7 & 29.5 & 85.2 & 26.4 & 15.6 & 76.7 & 91.7 & \textbf{31.0} & 12.5 & 71.5 & 41.1 & 47.3 & 27.7 & 49.1 \\
    & MaxSquare~\cite{chen2019maxsquare} & 81.2 & 32.8 & 85.4 & \textbf{31.9} & 14.7 & \textbf{78.3} & \textbf{92.7} & 28.3 & 8.6 & 68.2 & \textbf{42.2} & 51.3 & \textbf{32.4} & 49.8 \\
    % \cline{2-16}
    & Source Only (our) & 82.9 & 33.4 & 86.1 & 22.7 & 15.1 & 76.4 & 92.2 & 15.1 & 17.2 & 70.9 & 38.7 & 48.9 & 17.4 & 47.5  \\
	& ASA (our) & 82.6 & 30.8 & \textbf{86.3} & 26.0 & 15.1 & 77.0 & 92.3 & 28.4 & 9.1 & 71.9 & 39.8 & 47.4 & 31.0 & 49.1 \\
	\toprule
	\end{tabular}
 }
\end{table*}

\begin{figure*}
  \centering
  \includegraphics[width=1.0\linewidth]{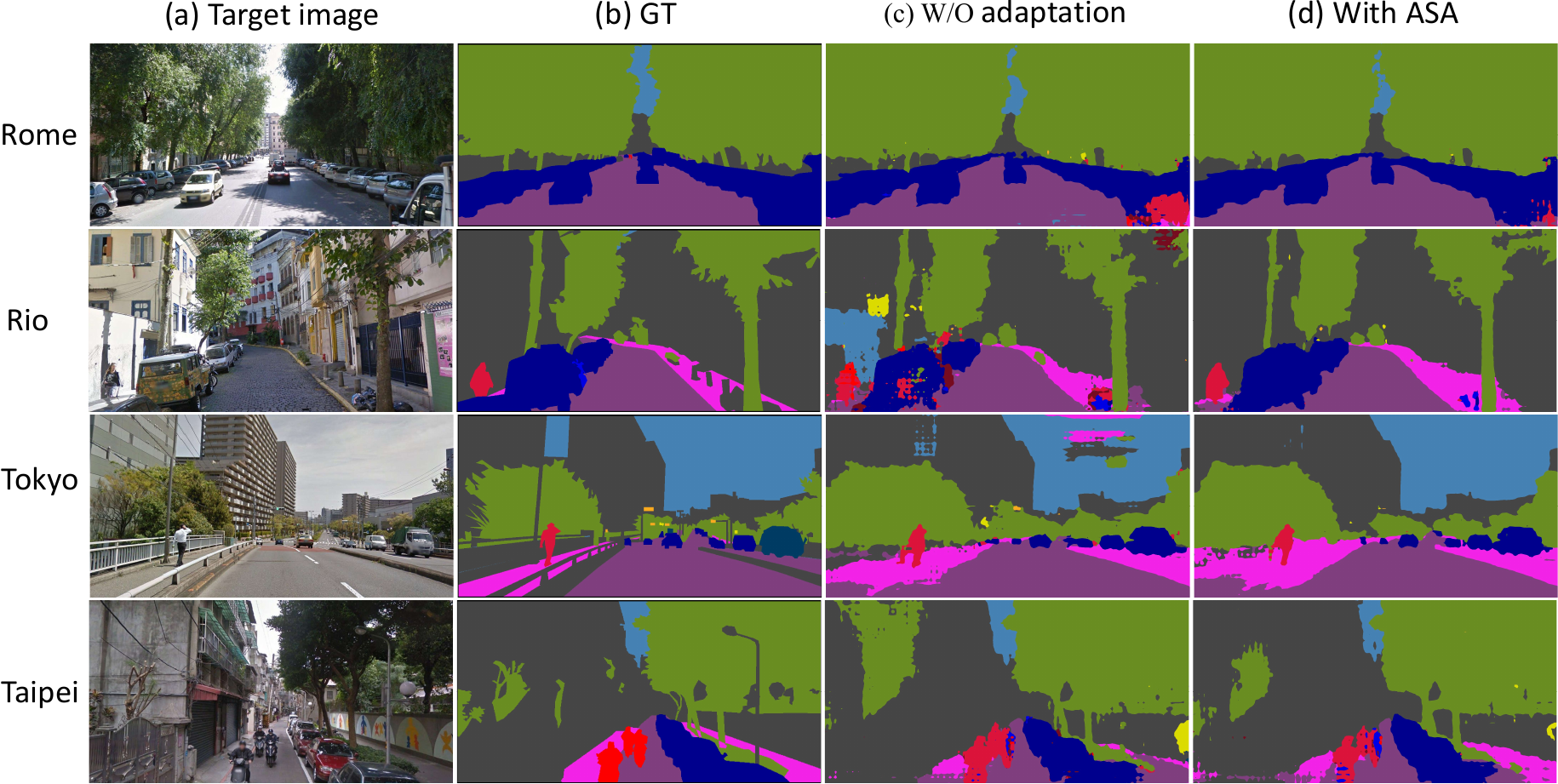}
  \caption{Some qualitative results on Cityscapes $\rightarrow$ Cross-city adaptation. From left to right: (a) target image; (b) ground-truth segmentation; (c) segmentation without adaptation; (d) segmentation with affinity space alignment (ASA) adaptation.}
  \label{fig:crosscity}
\end{figure*}

\smallskip
\textit{Image translation and self-training} may further be complementary to distribution alignment based methods~\cite{li2019BDL, tsai2019sodpr}.
For a fair comparison with BDL~\cite{li2019BDL} and SODPR~\cite{tsai2019sodpr},
we also equip the proposed ASA with both image translation and self-training strategies to further boost performance.
As depicted in Tab.~\ref{table:sota_rst}, we achieve the state-of-the-art performance on both GTA5 $\rightarrow$ Cityscapes and SYNTHIA $\rightarrow$ Cityscapes benchmarks, which further demonstrates the complementarity of ASA to image translation and self-training.
Specifically, we get 49.1\% mIoU on GTA5 $\rightarrow$ Cityscapes adaptation, outperforming BDL~\cite{li2019BDL} by 0.6\% mIoU and SODPR~\cite{tsai2019sodpr} by 2.6\% mIoU, respectively. On SYNTHIA $\rightarrow$ Cityscapes adaptation, we achieve 53.8 \% mIoU, which improves BDL~\cite{li2019BDL} by 2.4\% mIoU and SODPR~\cite{tsai2019sodpr} by 7.3 \% mIoU, respectively.

\begin{table}
      \small
      \caption{Quantitative results in terms of Dice coefficient for adapting REFUGE to RIM-ONE-r3 on `test' split of RIM-ONE-r3.}
      \centering
      \resizebox{0.45\textwidth}{!}{
      \begin{tabular}{c|c|c|c}
        \bottomrule
        \multirow{2}{*}{Method} & \multicolumn{3}{c}{RIM-ONE-r3} \\
        \cline{2-4} & $Dice_{cup}$  & $Dice_{disc}$ &$Mean$ \\ \hline
        Source only& {0.751} &{0.793} & 0.772 \\ \hline
    	Oracle & 0.856 & 0.968 &0.914\\ \hline \hline
    	AdaptSegNet~\cite{tsai2018adaptsegnet} & 0.775 & 0.893 & 0.834\\ \hline
    	\textbf{ASA} (ours) & \textbf{0.806}& \textbf{0.900} & \textbf{0.853}\\
        \toprule

       \end{tabular}
       }
        \label{table:med}
\end{table}

\subsection{Ablation Study}
\label{sec:ablation_study}
We conduct ablation study on two major components of the proposed method: 1) loss weight hyper-parameters $\lambda_{ASC}$ in Eq.~\eqref{eq:asctotalloss} and $\lambda_{ASA}$ in Eq.~\eqref{eq:asaloss}; 2) Different connectivity in defining the affinity space.

\medskip

\noindent\textbf{Ablation study on hyper-parameters.}
The weighting factor $\lambda_{ASC}$ for ASC and $\lambda_{ASA}$ for ASA are two important hyper-parameters. We adopt a sensitivity analysis
on these two hyper-parameters.
We evaluate the performance of both ASC and ASA with different weighting factors on GTA5$\rightarrow$ Cityscapes adaptation. As depicted in Tab.~\ref{tab:gta5_para}, both ASC and ASA can tolerate a large range of weighting factors. And we set both $\lambda_{ASC}$ and $\lambda_{ASA}$ to 0.001 in all experiments.

\medskip
\noindent\textbf{Ablation study on different connectivity.} We also study the effect of different connectivity settings involved in defining the affinity space. For that, we conduct experiments with 4-connectivity and 8-connectivity on the GTA5 $\rightarrow$ Cityscapes adaptation. As depicted in Tab.~\ref{table:ablation}, both ASC and ASA strategies for affinity space adaptation achieve better performance with 8-connectivity, showing that the co-occurring patterns revealing the output structure is important for unsupervised domain adaptation in semantic segmentation.
To further understand the impact of different connectivity, we also visualize the diagonal affinity maps for ASA with different connectivity settings on GTA5 $\rightarrow$ Cityscapes adaptation on some target images. As shown in Fig.~\ref{fig:N_ablation}, for each pixel, we calculate the average Cosine similarity with its 4 diagonal neighbors (see red arrows on the bottom left of target images in Fig.~\ref{fig:N_ablation}). It can be observed that ASA with 8-connectivity (C8) setting achieves better diagonal affinity than ASA under 4-connectivity (C4) setting, which leads to better segmentation performance.

\subsection{Model complexity analysis}
\label{subsec:complexityanalysis}
We also analyze the amount of computation and memory footprint during training.
The extra computational complexity of both ASC and ASA is $\mathcal{O}(N \times H \times W)$, where $H \times W$ donates the spatial resolution of output prediction and $N$ denotes the number of involved adjacent pixels.
As shown in Tab.~\ref{table:ablation}, compared with AdaptSegNet~\cite{tsai2018adaptsegnet}, both the proposed ASC and ASA achieve better performance while slightly increasing the computation cost and memory usage during training.

\subsection{Real-to-Real adaptation results}

Based on the synthetic-to-real adaptation evaluation, the proposed ASA strategy is in general more effective than the proposed ASC strategy for affinity space adaptation. We evaluate the proposed ASA on two real-to-real segmentation adaptations to further demonstrate the effectiveness of the proposed method.

\medskip

\noindent\textbf{Cross-City $\rightarrow$ Cityscapes.}
Most existing works for UDA semantic segmentation mainly focus on synthetic-to-real situation, especially on synthetic-to-cityscapes task. In practice, domain shift across different cities is a more realistic and challenging scenario. To verify the effectiveness of the proposed method on real-to-real adaptation, we conduct experiments on adapting semantic segmentation from Cityscapes dataset to Cross-city dataset.
We list our baseline and ASA results in Tab.~\ref{table:cross_city}, and compare with some other state-of-the-art methods.
As shown in Tab.~\ref{table:cross_city}, for small domain gap in real-to-real scenario, the proposed approach also achieves consistent improvement for different cities. Some qualitative segmentation results are illustrated in Fig.~\ref{fig:crosscity}. The proposed ASA effectively improves the segmentation results without any adaptation.

\begin{figure}
    \centering
    \includegraphics[width=1.0\linewidth]{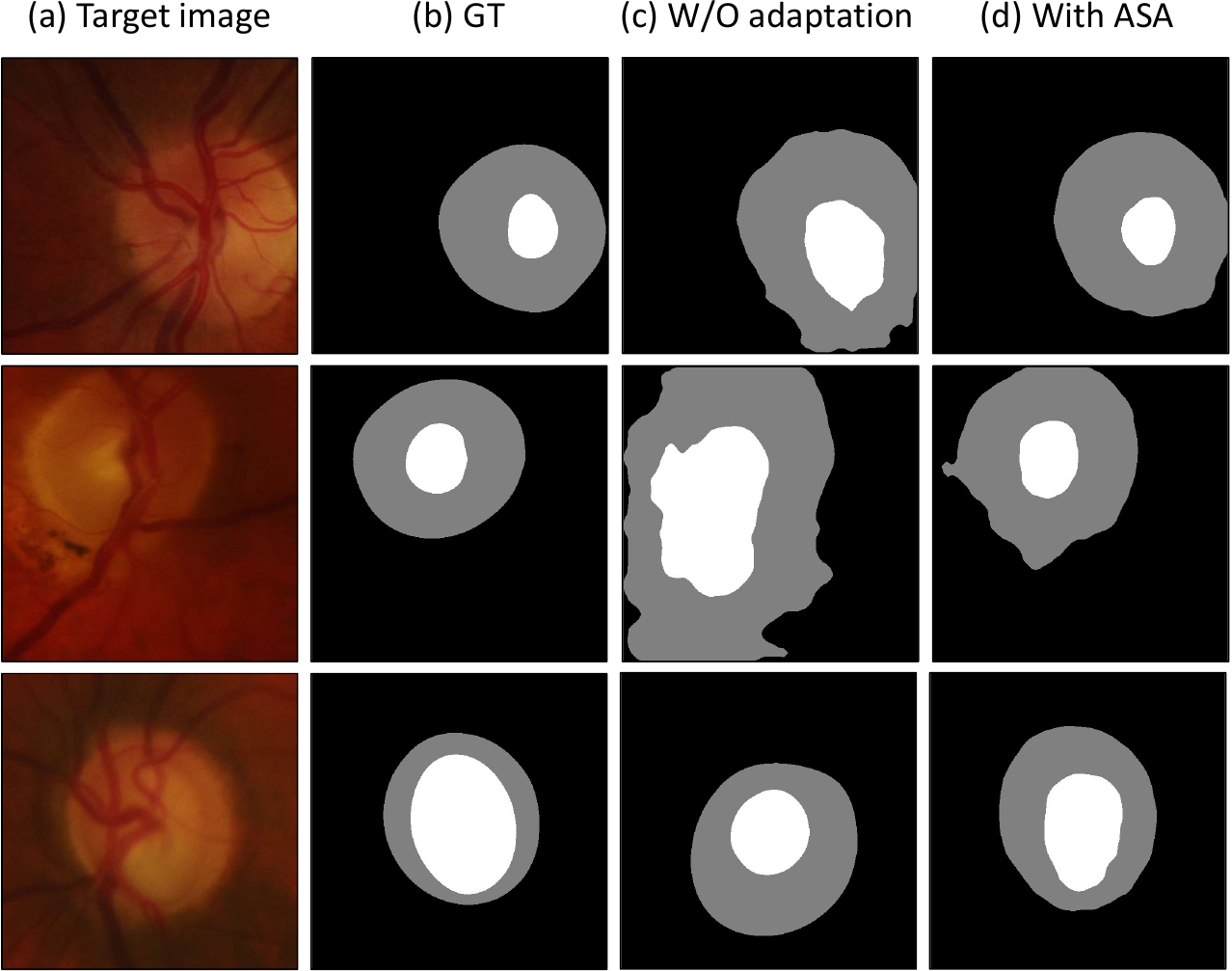}
    \caption{Some qualitative results on REFUGE $\rightarrow$ RIM-ONE-r3 adaptation.}
    \label{fig:med}
\end{figure}

\medskip
\noindent\textbf{REFUGE $\rightarrow$ RIM-ONE-r3.}
In addition to urban scene adaptation, the domain gap is also an important issue in medical image analysis. To verify the generalizability of the proposed approach, we conduct an experiment on unsupervised domain adaptation for retinal optic disk and cup segmentation.
Specifically, we adopt Deeplabv3-plus~\cite{deeplabv3plus2018} as the baseline, and evaluate the most related AdaptSegNet~\cite{tsai2018adaptsegnet} and the proposed ASA using the same experimental settings.
As shown in Tab.~\ref{table:med}. The proposed ASA consistently outperforms the baseline and AdaptSegNet~\cite{tsai2018adaptsegnet}. Some qualitative results are illustrated in Fig.~\ref{fig:med}. The proposed ASA effectively bridges up the domain gap between the source and target domains.

\begin{figure}
    \centering
    \includegraphics[width=1.0\linewidth]{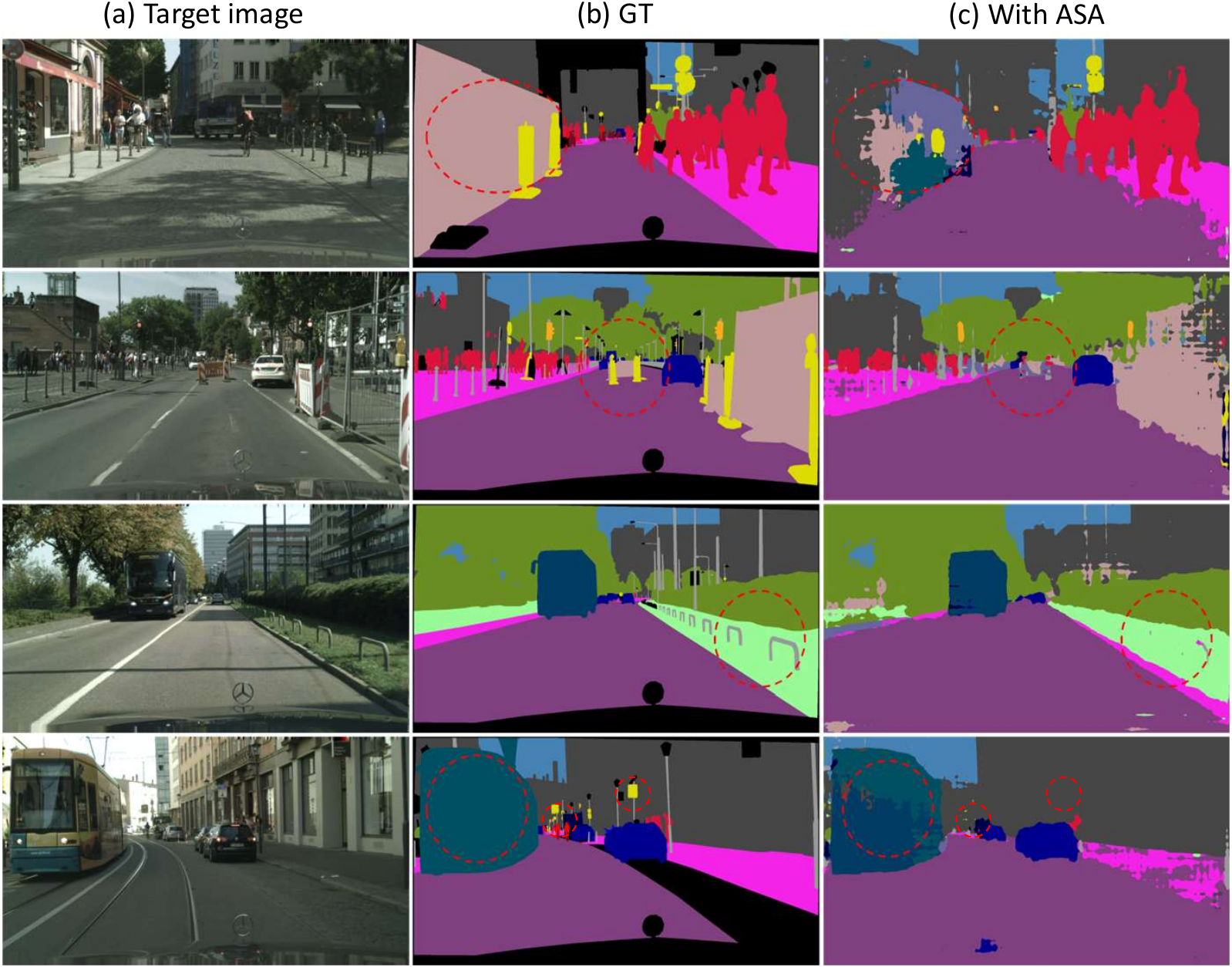}
    \caption{Some failure examples (see the region enclosed by red circles) from GTA5 $\rightarrow$ Cityscapes adaptation.}
    \label{fig:failure}
\end{figure}

\subsection{Weakness}
As demonstrated in previous experimental results, the proposed method performs well in most situations. However, it stills fails to handle some difficult problems, such as delicate structure, region inconsistency.
The proposed method also works less well for lower frequent classes (\eg, light and pole).
Some failure cases are shown in Fig.~\ref{fig:failure}.
Note that these difficult problems are also challenging for the other state-of-the-art methods.

\section{Conclusion}
\label{sec:conclusion}

In this paper, we address the problem of unsupervised domain adaptation for semantic segmentation. Considering that the output of semantic segmentation is usually structured and has invariant structures across domains, we propose to leverage such invariance through exploiting affinity relationship between adjacent pixels in the output level instead of adapting domains on individual per-pixel information for most state-of-the-art methods. To this end, we introduce the concept of affinity space encoding the affinity relationship. We exploit two affinity space adaptation strategies: affinity space cleaning and adversarial affinity space alignment. Both affinity space adaptation schemes effectively align the co-occurring patterns that reveal the output structure of semantic segmentation of source and target domain, leading to superior performance over some state-of-the-art methods. This demonstrates for the first time that the affinity relationship is beneficial for unsupervised domain adaptation in semantic segmentation.
In the future, we plan to investigate the combination of the two proposed affinity space adaptation strategies and combine the proposed output level adaptation with other adaptations on image and/or feature level. We would also like to explore the complementary of the proposed method with recent class- or region-conditioned adaptation to further boost the performance.

\section*{Acknowledgment}
This work was supported in part by the Major Project for New Generation of AI under Grant no. 2018AAA0100400,
NSFC 61703171, and NSF of Hubei Province of China
under Grant 2018CFB199.
Dr. Yongchao Xu was supported by
the Young Elite Scientists Sponsorship Program by CAST.

%\section*{Acknowledgment}

\ifCLASSOPTIONcaptionsoff
  \newpage
\fi



%\bibliographystyle{IEEEtran}


% \end{thebibliography}

\bibliographystyle{IEEEtran}
\bibliography{ref.bib}
\end{document}